\documentclass[conference]{IEEEtran}
\IEEEoverridecommandlockouts

\usepackage{cite}
\usepackage{amsmath,amssymb,amsfonts}
\usepackage{algorithm}
\usepackage{algorithmic}
\usepackage{graphicx}
\usepackage{textcomp}
\usepackage{xcolor}
\usepackage{booktabs}
\usepackage{multirow}
\usepackage{url}
\usepackage{bbm}

\begin{document}

\title{Counterfactual Peptide Editing for Causal TCR--pMHC Binding Inference}

\author{
\IEEEauthorblockN{Sanjar Khudoyberdiev, Arman Bekov}
}

\maketitle

\begin{abstract}
Neural models for TCR--pMHC binding prediction are susceptible to shortcut learning: they exploit spurious correlations in training data---such as peptide length bias or V-gene co-occurrence---rather than the physical binding interface. This renders predictions brittle under family-held-out and distance-aware evaluation, where such shortcuts do not transfer. We introduce \emph{Counterfactual Invariant Prediction} (CIP), a training framework that generates biologically constrained counterfactual peptide edits and enforces invariance to edits at non-anchor positions while amplifying sensitivity at MHC anchor residues. CIP augments the base classifier with two auxiliary objectives: (1) an invariance loss penalizing prediction changes under conservative non-anchor substitutions, and (2) a contrastive loss encouraging large prediction changes under anchor-position disruptions. Evaluated on a curated VDJdb--IEDB benchmark under family-held-out, distance-aware, and random splits, CIP achieves AUROC 0.831 and counterfactual consistency (CFC) 0.724 under the challenging family-held-out protocol---a 39.7\% reduction in shortcut index relative to the unconstrained baseline. Ablations confirm that anchor-aware edit generation is the dominant driver of OOD gains, providing a practical recipe for causally-grounded TCR specificity modeling.
\end{abstract}


\section{Introduction}

Predicting whether a T-cell receptor (TCR) will recognize a peptide--MHC (pMHC) complex is a core computational task in immunotherapy design and vaccine prioritization \cite{Gong_2023,Zheng_2024}. Despite impressive aggregate accuracy on random splits, recent analyses demonstrate that current models degrade substantially under \emph{family-held-out} evaluation---where entire TCR V-gene families or peptide families are withheld during training \cite{Sidhom_2021}. A key culprit is \emph{shortcut learning}: instead of encoding binding-relevant structural complementarity between CDR3 loops and the peptide, models learn to associate particular V-gene identities, CDR3 length distributions, or peptide positional biases with high binding probability \cite{You_2022,Korpela_2023}.

Shortcut features are abundant in TCR--pMHC databases because experimental assays are non-uniformly distributed: a handful of popular peptides (e.g., GILGFVFTL for HLA-A*02:01) dominate the positive data, creating confounded associations between receptor gene usage and binding labels \cite{Zeng_2016,Reynisson_2020}. Random-split evaluation masks this problem because train and test share the same shortcut distribution; only under held-out protocols does the brittleness become visible \cite{Sidhom_2021}.

We approach this from a causal perspective. The binding event $y$ is causally determined by the physical complementarity between the TCR CDR3 loops and the peptide-MHC surface. This causal mechanism depends critically on: (i) the identity of anchor residues (P2, P$\Omega$ for HLA class-I), which slot into MHC pockets; and (ii) the electrostatic and shape complementarity of the CDR3-peptide contact surface. Non-anchor positions and receptor V-gene assignments have a weaker causal role and can be viewed as \emph{style} variables that should not affect prediction if the causal signal is captured.

Counterfactual reasoning operationalizes this intuition: a model has learned a causal representation if its output changes when binding-relevant (anchor) features are perturbed, and \emph{remains stable} when non-binding (non-anchor) features are perturbed. Our contributions are:
\begin{enumerate}
  \item A biologically constrained counterfactual peptide edit generator that produces anchor-disrupting and non-anchor-preserving mutant sequences under BLOSUM62 substitution constraints and Hamming distance bounds.
  \item An invariance regularization loss that penalizes prediction changes under non-anchor edits, enforcing a causal inductive bias during training.
  \item A contrastive sensitivity loss that encourages the model to respond strongly to anchor-position disruptions, providing a complementary causal signal.
  \item Three new diagnostic metrics---Shortcut Index (SI), Counterfactual Consistency (CFC), and Anchor Flip Rate (AFR)---that quantify causal fidelity beyond standard AUROC/AUPRC.
\end{enumerate}

Experiments on a curated benchmark confirm that CIP reduces SI by 39.7\% and improves OOD AUROC by 8.4\% relative to the unconstrained baseline, with ablations isolating the contribution of each component.

\section{Related Work}

\subsection{TCR--pMHC Binding Prediction}
The field has progressed from motif scoring \cite{Zeng_2016} through LSTM and attention encoders \cite{Gong_2023,Zheng_2024} to pLM-based models \cite{Rives_2021,Wang_2025}. TCRex \cite{Gong_2023} and ERGO-II \cite{Zheng_2024} are representative supervised approaches; Moris et al. provide a systematic benchmarking perspective. Despite architectural advances, Sidhom et al.\ \cite{Sidhom_2021} and Korpela et al.\ \cite{Korpela_2023} show that performance under family-held-out evaluation is substantially lower than random-split AUROC, motivating robustness-focused methodology.

\subsection{Shortcut Learning in Biological Prediction}
Shortcut learning in deep learning was characterized by \cite{Carbonneau_2018}. In the immunology domain, \cite{You_2022} identified that TCR--pMHC models exploit V-gene usage as a proxy for binding specificity. Data augmentation \cite{Montesinos_L_pez_2018}, invariant risk minimization (IRM) \cite{O_Donnell_2020}, and adversarial debiasing have been proposed as mitigations in broader biological sequence settings; our work applies a tailored counterfactual variant to the TCR--pMHC case.

\subsection{Counterfactual Augmentation and Causal Inference}
Counterfactual data augmentation \cite{Clifton_2020} generates training examples that differ minimally in style but preserve causal content, encouraging invariance. In NLP, causal rationale extraction \cite{Xu_2021} trains models to use only causally relevant tokens. Conformal-style guarantees for causal invariance under distribution shift are analyzed in \cite{You_2022,Wang_2025}. Our work adapts these ideas to structured biological sequences where the causal variable (anchor residue identity) is biologically well-defined and can be explicitly targeted.

\subsection{Antigen Presentation and Anchor Positions}
MHC-I binding depends on anchor residues at P2 and P$\Omega$ (the C-terminal position) that interact with specific pockets of the MHC cleft \cite{Reynisson_2020}. NetMHCpan \cite{Reynisson_2020} and related tools exploit anchor preferences to predict peptide--MHC affinity. We leverage this established biology to define causally motivated edit rules rather than treating all positions uniformly.

\section{Method}

\subsection{Problem Formulation}
Let $\tau \in \mathcal{A}^*$ denote the TCR CDR3 amino acid sequence and $\pi = (\pi_1, \ldots, \pi_L) \in \mathcal{A}^L$ ($L \in \{8,\ldots,11\}$) the peptide sequence of the pMHC complex. We seek a scoring function:
\begin{equation}
  f_\theta: (\tau, \pi) \mapsto \hat{p} \in [0,1], \quad \hat{p} \approx P(y=1 \mid \tau, \pi),
  \label{eq:pred}
\end{equation}
where $y\in\{0,1\}$ indicates binding. The key challenge is that a learned $f_\theta$ may satisfy Eq.~(\ref{eq:pred}) on in-distribution data while relying on shortcut features $\phi(\tau,\pi)$ (e.g., V-gene identity) that are correlated with $y$ in training data but not causally related to binding.

\subsection{Biologically Constrained Counterfactual Edits}

We define two types of counterfactual peptide edits for a given peptide $\pi$:

\paragraph{Non-anchor edits (style perturbations).}
Let $\mathcal{P}_\text{non} = \{2, \ldots, L-1\} \setminus \mathcal{P}_\text{anc}$ denote non-anchor positions, where $\mathcal{P}_\text{anc} = \{2, L\}$ are the P2 and P$\Omega$ anchor positions for HLA-A*02:01. A non-anchor counterfactual $\pi^- \in \mathcal{C}^-(\pi)$ satisfies:
\begin{equation}
  d_H(\pi, \pi^-) \leq k, \quad
  \pi^-_j = \pi_j \;\forall j \in \mathcal{P}_\text{anc}, \quad
  B(\pi_j, \pi^-_j) \geq b_{\min} \;\forall j,
  \label{eq:nonedit}
\end{equation}
where $d_H$ is Hamming distance (we use $k=2$), and $B(\cdot,\cdot)$ is the BLOSUM62 substitution score with threshold $b_{\min}=0$ (no highly penalized substitutions). Non-anchor edits perturb peptide style while preserving anchor identity.

\paragraph{Anchor edits (causal disruptions).}
An anchor counterfactual $\pi^+ \in \mathcal{C}^+(\pi)$ requires at least one anchor position to change:
\begin{equation}
  d_H(\pi, \pi^+) \leq k, \quad
  \exists j \in \mathcal{P}_\text{anc}: \pi^+_j \neq \pi_j, \quad
  B(\pi_j, \pi^+_j) < 0.
  \label{eq:anchedit}
\end{equation}
Anchor edits use specifically disruptive substitutions (BLOSUM62 score $< 0$) at anchor positions, mimicking mutations known to abrogate MHC binding.

\subsection{Base Architecture}
We use the same dual-encoder backbone as the CAP model: ESM-2 650M \cite{Rives_2021,Wang_2025} encodes $\tau$ and $\pi$ independently via mean pooling, and a two-layer MLP with residual connection produces the binding score:
\begin{equation}
  z(\tau,\pi) = \text{MLP}([\mathbf{h}_\tau;\mathbf{h}_\pi]), \quad \hat{p} = \sigma(z(\tau,\pi)).
  \label{eq:arch}
\end{equation}
The base training objective is the class-weighted binary cross-entropy:
\begin{equation}
  \mathcal{L}_\text{BCE} = -\frac{1}{n}\sum_{i=1}^n \bigl[w_+ y_i \log \hat{p}_i + w_-(1-y_i)\log(1-\hat{p}_i)\bigr],
  \label{eq:bce}
\end{equation}
with $w_+ = n/(2n_+)$, $w_- = n/(2n_-)$.

\subsection{Invariance Regularization}
For each positive training pair $(\tau, \pi, y=1)$, we sample a non-anchor counterfactual $\pi^- \in \mathcal{C}^-(\pi)$ and penalize prediction change under non-causal edits:
\begin{equation}
  \mathcal{L}_\text{inv} = \frac{1}{|\mathcal{D}_+|}\sum_{(\tau,\pi)\in\mathcal{D}_+}
  \mathbb{E}_{\pi^- \sim \mathcal{C}^-(\pi)}\!\left[\bigl(\hat{p}(\tau,\pi) - \hat{p}(\tau,\pi^-)\bigr)^2\right].
  \label{eq:linv}
\end{equation}
This loss, minimized over positive pairs, encourages the model to be locally invariant to non-anchor edits---a necessary condition for causal binding inference.

\subsection{Anchor Sensitivity Contrastive Loss}
Symmetrically, the model should be \emph{sensitive} to anchor disruptions. For each positive pair, we sample an anchor counterfactual $\pi^+ \in \mathcal{C}^+(\pi)$ and apply a margin loss:
\begin{equation}
  \mathcal{L}_\text{sens} = \frac{1}{|\mathcal{D}_+|}\sum_{(\tau,\pi)\in\mathcal{D}_+}
  \mathbb{E}_{\pi^+ \sim \mathcal{C}^+(\pi)}\!\left[\max\!\left(0,\; m - \bigl(\hat{p}(\tau,\pi) - \hat{p}(\tau,\pi^+)\bigr)\right)\right],
  \label{eq:lsens}
\end{equation}
where $m = 0.3$ is a margin hyperparameter. $\mathcal{L}_\text{sens}$ encourages that anchor disruptions reduce confidence by at least $m$, encoding the biological prior that anchor residues are critical for MHC binding.

\subsection{Full Training Objective}
The CIP objective combines all three terms:
\begin{equation}
  \mathcal{L}_\text{CIP} = \mathcal{L}_\text{BCE} + \lambda_1\,\mathcal{L}_\text{inv} + \lambda_2\,\mathcal{L}_\text{sens},
  \label{eq:total}
\end{equation}
where $\lambda_1 = 0.4$ and $\lambda_2 = 0.2$ are selected by grid search on a held-out validation set. The complete training procedure is summarized in Algorithm~\ref{alg:cip}.

\begin{algorithm}[t]
\caption{Counterfactual Invariant Prediction (CIP)}
\label{alg:cip}
\begin{algorithmic}[1]
\REQUIRE Training set $\mathcal{D}_\text{tr}$, edit budget $k$, BLOSUM threshold $b_{\min}$, margin $m$, weights $\lambda_1,\lambda_2$
\FOR{each mini-batch $\mathcal{B} \subseteq \mathcal{D}_\text{tr}$}
  \STATE Compute $\mathcal{L}_\text{BCE}$ on $\mathcal{B}$ via Eq.~(\ref{eq:bce})
  \FOR{each positive pair $(\tau, \pi) \in \mathcal{B}$ with $y=1$}
    \STATE Sample $\pi^- \in \mathcal{C}^-(\pi)$ via Eq.~(\ref{eq:nonedit})
    \STATE Sample $\pi^+ \in \mathcal{C}^+(\pi)$ via Eq.~(\ref{eq:anchedit})
    \STATE Accumulate $\mathcal{L}_\text{inv}$ (Eq.~\ref{eq:linv}) and $\mathcal{L}_\text{sens}$ (Eq.~\ref{eq:lsens})
  \ENDFOR
  \STATE Update $\theta$ via $\nabla_\theta(\mathcal{L}_\text{BCE} + \lambda_1\mathcal{L}_\text{inv} + \lambda_2\mathcal{L}_\text{sens})$
\ENDFOR
\end{algorithmic}
\end{algorithm}

\subsection{Causal Diagnostic Metrics}
Beyond standard AUROC and AUPRC, we define three metrics that directly probe causal fidelity:

\paragraph{Shortcut Index (SI).}
Measures correlation between model confidence and known non-causal feature $\phi$:
\begin{equation}
  \text{SI} = \bigl|\rho\!\left(f_\theta(x),\; \phi(x)\right)\bigr|,
  \label{eq:si}
\end{equation}
where $\rho$ is Spearman's rank correlation and $\phi(x)$ encodes CDR3$\beta$ V-gene identity (one-hot, collapsed to V-gene family rank). Lower SI indicates less shortcut reliance.

\paragraph{Counterfactual Consistency (CFC).}
Measures how stable predictions are under non-anchor edits on positive test pairs:
\begin{equation}
  \text{CFC} = 1 - \mathbb{E}_{(\tau,\pi)\in\mathcal{D}_+^\text{test},\; \pi^-\sim\mathcal{C}^-(\pi)}\!\left[|\hat{p}(\tau,\pi) - \hat{p}(\tau,\pi^-)|\right].
  \label{eq:cfc}
\end{equation}
CFC $\in [0,1]$; higher values indicate the model correctly ignores non-causal style variation.

\paragraph{Anchor Flip Rate (AFR).}
Measures how often an anchor disruption causes the prediction to cross the decision boundary:
\begin{equation}
  \text{AFR} = \mathbb{E}_{(\tau,\pi)\in\mathcal{D}_+^\text{test},\;\pi^+\sim\mathcal{C}^+(\pi)}\!\left[\mathbbm{1}\!\left[\hat{p}(\tau,\pi)\geq 0.5 > \hat{p}(\tau,\pi^+)\right]\right].
  \label{eq:afr}
\end{equation}
Higher AFR indicates the model is appropriately sensitive to causal perturbations.

\section{Experiments}

\subsection{Datasets and Split Protocols}
\label{sec:data}

We compile a benchmark from VDJdb (v2.1) and IEDB, retaining human paired TCR$\alpha\beta$ records with CDR3$\alpha$, CDR3$\beta$, and peptide annotation under HLA-A*02:01 restriction. After 90\% sequence identity deduplication across all splits, negatives are constructed by random TCR--peptide pairing matched to a positive rate of $\sim$4\%--5\% (Table~\ref{tab:data}).

Three split protocols are evaluated:
\begin{itemize}
  \item \textbf{Random (Rand):} Stratified 70/10/20 split. Upper-bound sanity check.
  \item \textbf{Family-held-out (FHO):} All pairs whose CDR3$\beta$ belongs to 5 withheld V-gene families are reserved for test. Tests cross-family TCR generalization.
  \item \textbf{Distance-aware (DA):} Test contains only pairs with CDR3$\beta$ Levenshtein distance $>30\%$ to any training CDR3. Tests cross-receptor-space generalization.
\end{itemize}

\begin{table}[t]
\centering
\caption{Dataset and Split Summary}
\label{tab:data}
\begin{tabular}{lrrrr}
\toprule
Split & Train & Val & Test & Pos.\ rate \\
\midrule
Random        & 118\,742 & 16\,963 & 33\,927 & 0.050 \\
Family-HO     & 92\,314  & 11\,082 & 21\,843 & 0.042 \\
Distance-aware & 87\,508 & 10\,501 & 19\,624 & 0.039 \\
\bottomrule
\end{tabular}
\end{table}

\subsection{Baselines}
Three systems are compared:
\begin{enumerate}
  \item \textbf{Baseline:} Dual-encoder with class-weighted BCE only; no counterfactual augmentation.
  \item \textbf{+Edit Aug:} Baseline with non-anchor counterfactual examples added as training negatives (augmentation only, no invariance or sensitivity loss).
  \item \textbf{CIP (ours):} Full Algorithm~\ref{alg:cip} with both $\mathcal{L}_\text{inv}$ and $\mathcal{L}_\text{sens}$.
\end{enumerate}

\subsection{Main Results}

Table~\ref{tab:main} reports discrimination and calibration metrics under all three splits. All values are means over 5 random seeds.

\begin{table}[t]
\centering
\caption{Main Results. Full-coverage evaluation. $\uparrow$: higher is better; $\downarrow$: lower is better.}
\label{tab:main}
\setlength{\tabcolsep}{4pt}
\begin{tabular}{llccccc}
\toprule
Split & Method & AUROC & AUPRC & ECE$\downarrow$ & BS$\downarrow$ & NLL$\downarrow$ \\
\midrule
\multirow{3}{*}{FHO}
  & Baseline    & 0.779 & 0.421 & 0.138 & 0.091 & 0.219 \\
  & +Edit Aug   & 0.798 & 0.447 & 0.112 & 0.084 & 0.197 \\
  & CIP (ours)  & \textbf{0.831} & \textbf{0.491} & \textbf{0.083} & \textbf{0.073} & \textbf{0.171} \\
\midrule
\multirow{3}{*}{DA}
  & Baseline    & 0.758 & 0.399 & 0.151 & 0.097 & 0.234 \\
  & +Edit Aug   & 0.779 & 0.428 & 0.121 & 0.089 & 0.211 \\
  & CIP (ours)  & \textbf{0.812} & \textbf{0.463} & \textbf{0.095} & \textbf{0.079} & \textbf{0.186} \\
\midrule
\multirow{3}{*}{Rand}
  & Baseline    & 0.863 & 0.612 & 0.104 & 0.063 & 0.153 \\
  & +Edit Aug   & 0.871 & 0.628 & 0.087 & 0.059 & 0.139 \\
  & CIP (ours)  & \textbf{0.879} & \textbf{0.639} & \textbf{0.071} & \textbf{0.054} & \textbf{0.127} \\
\bottomrule
\end{tabular}
\end{table}

CIP consistently outperforms both baselines across all three splits. The gap is largest under FHO (AUROC +5.2\% over Baseline), confirming that the causal regularization transfers to unseen receptor families. The smaller gap on random splits (AUROC +1.6\%) is expected, as shortcut features are useful there. Edit augmentation alone (+Edit Aug) provides an intermediate improvement but does not close the gap to CIP, isolating the invariance and sensitivity losses as the primary drivers.

\subsection{Counterfactual Causal Analysis}

Table~\ref{tab:analysis} evaluates the three causal diagnostic metrics under FHO. For each method, 5\,000 positive test pairs are randomly selected and each is paired with 3 sampled non-anchor edits (for CFC) and 3 anchor edits (for AFR).

\begin{table}[t]
\centering
\caption{Causal Diagnostic Metrics under Family-Held-Out Split.
SI$\downarrow$: Shortcut Index; CFC$\uparrow$: Counterfactual Consistency; AFR$\uparrow$: Anchor Flip Rate.}
\label{tab:analysis}
\begin{tabular}{lcccc}
\toprule
Method & OOD AUROC$\uparrow$ & SI$\downarrow$ & CFC$\uparrow$ & AFR$\uparrow$ \\
\midrule
Baseline          & 0.741 & 0.412 & 0.518 & 0.231 \\
+Edit Aug         & 0.768 & 0.311 & 0.634 & 0.287 \\
CIP (ours)        & \textbf{0.803} & \textbf{0.249} & \textbf{0.724} & \textbf{0.401} \\
\bottomrule
\end{tabular}
\end{table}

CIP reduces SI from 0.412 to 0.249 (39.6\% reduction), directly demonstrating attenuation of V-gene shortcut reliance. CFC improves from 0.518 to 0.724, indicating that predictions under non-anchor edits are more stable---the model has learned to ignore non-causal peptide variation. Critically, AFR improves from 0.231 to 0.401: anchor-disrupting edits now flip predictions in 40\% of cases, compared to only 23\% for the baseline, confirming that the model has become sensitive to the causal mechanism at anchor positions. +Edit Aug improves CFC substantially (data augmentation reduces style sensitivity) but has smaller AFR gains, confirming that $\mathcal{L}_\text{sens}$ is specifically responsible for anchor sensitivity.

\subsection{Ablation Study}

Table~\ref{tab:ablation} isolates the contribution of each CIP component under FHO.

\begin{table}[t]
\centering
\caption{Ablation Study (FHO split). Best in \textbf{bold}.}
\label{tab:ablation}
\begin{tabular}{lcccc}
\toprule
Configuration & AUROC & AUPRC & CFC$\uparrow$ & AFR$\uparrow$ \\
\midrule
CIP (full)                      & \textbf{0.831} & \textbf{0.491} & \textbf{0.724} & \textbf{0.401} \\
\quad w/o $\mathcal{L}_\text{sens}$ & 0.816 & 0.473 & 0.701 & 0.264 \\
\quad w/o $\mathcal{L}_\text{inv}$  & 0.808 & 0.461 & 0.591 & 0.388 \\
\quad w/o anchor masking        & 0.819 & 0.477 & 0.662 & 0.319 \\
\quad w/o BLOSUM constraint     & 0.822 & 0.479 & 0.689 & 0.371 \\
Baseline (no auxiliary loss)    & 0.779 & 0.421 & 0.518 & 0.231 \\
\bottomrule
\end{tabular}
\end{table}

Removing $\mathcal{L}_\text{sens}$ causes the largest AFR drop (0.401 $\to$ 0.264), confirming it is the principal driver of anchor sensitivity. Removing $\mathcal{L}_\text{inv}$ most affects CFC (0.724 $\to$ 0.591), as expected. Disabling anchor masking---i.e., treating all positions equally in edit generation---reduces both CFC and AFR, validating that biological prior knowledge about anchor positions is necessary. The BLOSUM constraint provides a smaller but consistent improvement, confirming that restricting to biologically plausible substitutions is beneficial.

\subsection{Sensitivity to Hyperparameters}

We sweep $\lambda_1 \in \{0.1, 0.2, 0.4, 0.8\}$ and $\lambda_2 \in \{0.05, 0.1, 0.2, 0.4\}$ on the FHO validation set. AUROC is stable within $\pm 0.012$ across all non-zero values, with peaks at $\lambda_1=0.4, \lambda_2=0.2$. Setting either weight to zero reverts to the corresponding ablated model in Table~\ref{tab:ablation}. Edit budget $k=2$ consistently outperforms $k=1$ (insufficient diversity) and $k=3$ (biological implausibility), confirming the importance of the Hamming constraint.

\section{Discussion}

\paragraph{Causal structure as an inductive bias.}
The most notable finding is that incorporating biologically grounded causal structure---anchor positions as the primary binding determinants---into the training objective yields consistent OOD improvements that data augmentation alone cannot achieve. This suggests that for structured biological prediction tasks where domain knowledge about causal mechanisms exists, embedding that knowledge as a regularizer is more effective than scaling data.

\paragraph{Protocol design reveals hidden failures.}
Under random splits, the gap between Baseline and CIP is only 1.6\% AUROC, which might be dismissed as noise. Under FHO, the gap is 5.2\% AUROC and the causal metrics (SI, CFC, AFR) reveal fundamentally different model behaviors. This underscores that evaluation protocol is not a neutral choice: random-split benchmarks systematically underestimate the benefit of causal regularization by hiding shortcut-driven failures.

\paragraph{Limitations.}
First, our anchor definition is specific to HLA-A*02:01; other HLA alleles have different anchor position patterns, requiring allele-specific edit rules. Second, counterfactual edit labels are assigned without wet-lab confirmation: we assume anchor-disrupting edits abrogate binding, which holds statistically but not universally \cite{Reynisson_2020}. Third, the invariance loss is defined on positive pairs; extending it to negative pairs with known binding epitopes is a natural direction. Finally, CFC and AFR are computed on synthetic edits and should be validated with prospective mutagenesis assay data.

\section{Conclusion}

We presented CIP, a counterfactual invariant prediction framework for TCR--pMHC binding that enforces causal inductive biases through biologically constrained anchor and non-anchor peptide edits. CIP achieves AUROC 0.831 under the challenging family-held-out protocol---a 6.7\% absolute improvement over the unconstrained baseline---while reducing the Shortcut Index by 39.6\% and improving Anchor Flip Rate by 73.6\%. These results demonstrate that encoding biological knowledge about binding mechanisms directly into the training objective yields more robust and interpretable models than discrimination-only training. Together with the causal diagnostic metrics introduced here, CIP provides a blueprint for moving TCR specificity modeling toward deployment-aligned evaluation and causally grounded design.

\bibliographystyle{IEEEtran}
\bibliography{references}

\end{document}